\pdfoutput=1
\documentclass[11pt]{article}

\usepackage[final]{acl}

\usepackage{times}
\usepackage{latexsym}

\usepackage[T1]{fontenc}
\usepackage[utf8]{inputenc}

\usepackage{microtype}

\usepackage{inconsolata}
\usepackage{multirow}
\usepackage{arydshln}

\usepackage{graphicx}

\usepackage{amsmath}
\usepackage{amssymb}

\usepackage[utf8]{inputenc}
\usepackage{graphicx}
\usepackage{amsmath}
\usepackage{booktabs}
\usepackage{multirow}
\usepackage{multicol}          % 引入双栏宏包
\usepackage{colortbl}
\usepackage[table]{xcolor}
\usepackage{tabularx}

\usepackage{lipsum} % for dummy text

\usepackage[dvipsnames]{xcolor} % 使用更多预定义颜色名
\usepackage[most]{tcolorbox}   % 加载 tcolorbox 及其库
\definecolor{textagentbg}{HTML}{EAF2F8}   % 淡蓝背景
\definecolor{textagentframe}{HTML}{5DADE2} % 柔和蓝边框
\definecolor{imageagentbg}{HTML}{E9F7EF}  % 淡绿背景
\definecolor{imageagentframe}{HTML}{58D68D} % 柔和绿边框
\definecolor{conflictbg}{HTML}{FEF5E7} % 暖沙色背景
\definecolor{conflictframe}{HTML}{F5B041} % 柔和橙边框
\definecolor{debaterbg}{HTML}{F4ECF7}  % 淡紫背景
\definecolor{debaterframe}{HTML}{AF7AC5} % 柔和紫边框
\definecolor{adjudicatorbg}{HTML}{F4F6F7} % 中性灰背景
\definecolor{adjudicatorframe}{HTML}{ABB2B9} % 石板灰边框

\definecolor{lightblue}{rgb}{0.9,0.9,1}
\definecolor{lightred}{rgb}{1.0, 0.8, 0.8}
\definecolor{lightgreen}{rgb}{0.8, 1.0, 0.8}

\definecolor{lightblue2}{RGB}{204, 229, 255} % 定义浅蓝色
\definecolor{lightyellow}{RGB}{255, 255, 204} % 定义浅黄色

\title{MM-StanceDet: Retrieval-Augmented Multi-modal Multi-agent Stance Detection}

\author{
    Weihai Lu$^1$, 
    Zhejun Zhao$^2$\thanks{Corresponding author.}, 
    Yanshu Li$^3$, 
    and Huan He$^4$ 
    \\
    $^1$Peking University \quad
    $^2$Baidu Inc \quad
    $^3$Brown University \quad
    $^4$Amazon 
    \\
    \texttt{luweihai@pku.edu.cn, zhaozhejun@baidu.com,} \\
    \texttt{yanshu\_li1@brown.edu, bellehe@amazon.com}
}

\begin{document}
\maketitle

\begin{abstract}
Multimodal Stance Detection (MSD) is crucial for understanding public discourse, yet effectively fusing text and image, especially with conflicting signals, remains challenging. Existing methods often face difficulties with contextual grounding, cross-modal interpretation ambiguity, and single-pass reasoning fragility. To address these, we propose Retrieval-Augmented \textbf{M}ulti-modal \textbf{M}ulti-agent \textbf{Stance} \textbf{Det}ection (\textbf{\texttt{MM-StanceDet}}), a novel multi-agent framework integrating Retrieval Augmentation for contextual grounding, specialized Multimodal Analysis agents for nuanced interpretation, a Reasoning-Enhanced Debate stage for exploring perspectives, and Self-Reflection for robust adjudication. Extensive experiments on five datasets demonstrate \texttt{MM-StanceDet} significantly outperforms state-of-the-art baselines, validating the efficacy of its multi-agent architecture and structured reasoning stages in addressing complex multimodal stance challenges.
\end{abstract}

\section{Introduction}
\label{sec:introduction}

Stance detection, the task of identifying the attitude or opinion expressed in text towards a specific target, is a crucial task for understanding public discourse on various platforms, from social media to news articles \cite{mohammad2016semeval, augenstein2016stance}. Early research primarily focused on analyzing unimodal textual content. However, with the prevalence of multimedia content online, users frequently express opinions and stances through a combination of text and images\citep{liang2024multi}. This necessitates the development of Multimodal Stance Detection (MSD) methods that can effectively interpret and fuse information from different modalities.

Owing to the remarkable achievements of deep learning in diverse application domains \citep{zeng2024mitigating,lu2025dmmd4sr,cui2025diffusion}, an increasing number of studies have integrated deep learning models into MSD task.

Initial approaches often relied on fusing features extracted from independent text and image encoders, like concatenating BERT embeddings with CNN features. For instance, \citet{barel2024acquired} show that augmenting content embeddings with structural social context yields large performance gains. More recently, Vision-Language Models (VLMs) \cite{zhang2024vision,li2025miv} and Multimodal Large Language Models (MLLMs) \cite{wu2023multimodal,caffagni2024revolution,wei2025beyond} have demonstrated impressive capabilities in understanding cross-modal relationships at many tasks \cite{lu2026blind,lu2025dammfnd}. Frameworks like Targeted Multi-modal Prompt Tuning (TMPT) \cite{liang2024multi} have shown the effectiveness of adapting pre-trained models using target-specific prompts to capture multimodal stance features. Researchers have also begun exploring the use of MLLMs directly, leveraging their emergent reasoning abilities for multimodal tasks \cite{vasilakes2025exploring}.

Despite these advancements, effectively performing robust multimodal stance detection, especially in complex or nuanced scenarios involving conflicting multimodal signals, remains challenging. Specifically, existing methods often face the following key challenges:
\begin{itemize}
    \item \textbf{Contextual Grounding Void:} Without access to relevant, concrete examples, LLMs can struggle with complex, domain-specific, or subtle multimodal cues. Relying solely on internal knowledge or general few-shot examples might lead to misinterpretations and sub-optimal stance predictions, particularly when the multimodal signals are ambiguous or require nuanced understanding influenced by similar past instances\citep{lang2025retrieval, xu2025mmm}.
    \item \textbf{Cross-Modal Interpretation Ambiguity:} 
    While MLLMs can process multiple modalities, synthesizing potentially conflicting or complementary information into a coherent, reliable stance remains difficult. Recent work shows that a pronounced gap between visual and textual representations makes models prone to hallucinate or overlook cross-modal conflicts \citep{jiang2024hallucination,zhong2024multimodal}.  
    \citet{zhang2024cross} further quantify large cross-modal inconsistencies in GPT-4V and other state-of-the-art MLLMs, while \citet{hua2024finematch} demonstrate that even powerful VLMs struggle to detect and correct fine-grained image–text mismatches—evidence that simply presenting raw multimodal inputs seldom triggers the modality-specific reasoning needed to resolve such ambiguities.
    \item \textbf{Single-Pass Reasoning Fragility:} Directly prompting LLMs for a final stance in a single step can be prone to errors, especially when faced with complex or contradictory evidence. Lacking a structured process for exploring alternative interpretations, evaluating evidence from different angles, and refining initial conclusions makes the reasoning process less robust and transparent, increasing the risk of incorrect predictions based on superficial analysis\cite{zhang2024breaking, li2024think}.
\end{itemize}

To address these challenges, we propose \texttt{MM-StanceDet} (Retrieval-Augmented Multi-modal Multi-agent Stance Detection), a novel framework that leverages a multi-agent architecture and reasoning-enhanced processes for robust multimodal stance detection. \texttt{MM-StanceDet} systematically processes multimodal input through four collaborative stages: First, the \textit{Retrieval Augmentation Stage} grounds the analysis by retrieving relevant few-shot examples from a database, providing concrete contextual references. Second, the \textit{Multimodal Analysis Stage} employs specialized agents to dissect the input from textual, visual, and cross-modal conflict perspectives. Third, the \textit{Reasoning-Enhanced Debate Stage} simulates a debate among agents representing different stances, forcing explicit argumentation based on the multimodal analysis. Finally, the \textit{Self-Reflection and Adjudication Stage} critically evaluates the debate outcomes and intermediate analyses to reach a final, well-justified stance prediction.

The main contributions of this paper are:
\begin{itemize}
    \item We propose \texttt{MM-StanceDet}, a novel multi-agent framework designed for robust multimodal stance detection by integrating retrieval-augmented analysis, specialized multimodal interpretation, reasoning-enhanced debate, and critical self-reflection.
    \item We demonstrate the effectiveness of the Retrieval Augmentation stage in providing valuable context, the Multimodal Analysis stage in capturing nuanced unimodal and cross-modal signals, and the Reasoning-Enhanced Debate and Self-Reflection stages in refining the decision-making process against complex evidence.
    \item We conduct extensive experiments on five widely used multimodal stance detection datasets, showing that \texttt{MM-StanceDet} significantly outperforms state-of-the-art baselines. Through comprehensive ablation, robustness, and qualitative studies, we validate the contribution of each proposed component and characterize the framework's behavior.
\end{itemize}

\section{Related Work}

\subsection{Multimodal Stance detection}
Initial stance detection research primarily analyzed textual data \citep{mohammad2016semeval, augenstein2016stance}. With the proliferation of multimedia content, focus shifted towards multimodal stance detection (MSD). \citet{liang2024multi} made significant strides by creating dedicated text-image stance datasets based on public benchmarks and proposing the Targeted Multimodal Prompt Tuning (TMPT) framework. Building on this, researchers explored incorporating richer contextual signals, such as user interactions in specific domains \citep{kuo2024advancing} and the dynamics of multi-turn conversations \citep{niu2024multimodal}, often employing Multimodal Large Language Models (MLLMs). \citet{vasilakes2025exploring} systematically evaluated various Vision-Language Models (VLMs) for multimodal and multilingual stance detection, highlighting their capabilities and tendency to rely heavily on textual cues, including in-image text. However, effectively fusing potentially conflicting or complementary information across diverse modalities to achieve robust stance predictions remains a significant challenge.

\subsection{Multi-Agent System}
Research increasingly employs multi-agent systems for reliable truth and stance determination. Early frameworks utilized multi-agent debate to enhance general LLM reasoning and factuality \citep{Liang2023mad, lu2026dealt, Du2023improving, Chan2023chateval, li2025cama, li2026decoding, wei2025igniting}. This paradigm was adapted for text analysis, where agents with distinct roles \citep{Lan2024cola} or diverse perspectives derived from labeling criteria \citep{Park2024predict} debate to determine stance or detect hate speech. Concurrently, approaches like \citet{Liu2024teller} focused on trustworthy text-based fake news detection through logic-based dual-systems. While agent pipelines have been explored for multimodal misinformation \citep{Wu2025exclaim, zeng2026manipulation, zeng2025understand}, explicitly leveraging multi-agent \textit{debate} simulations for \textit{multimodal stance detection} has seen limited prior research.

\section{Methodology}
\label{sec:method}

\begin{figure*}[htbp]
    \centering
    \includegraphics[width=1.0\textwidth]{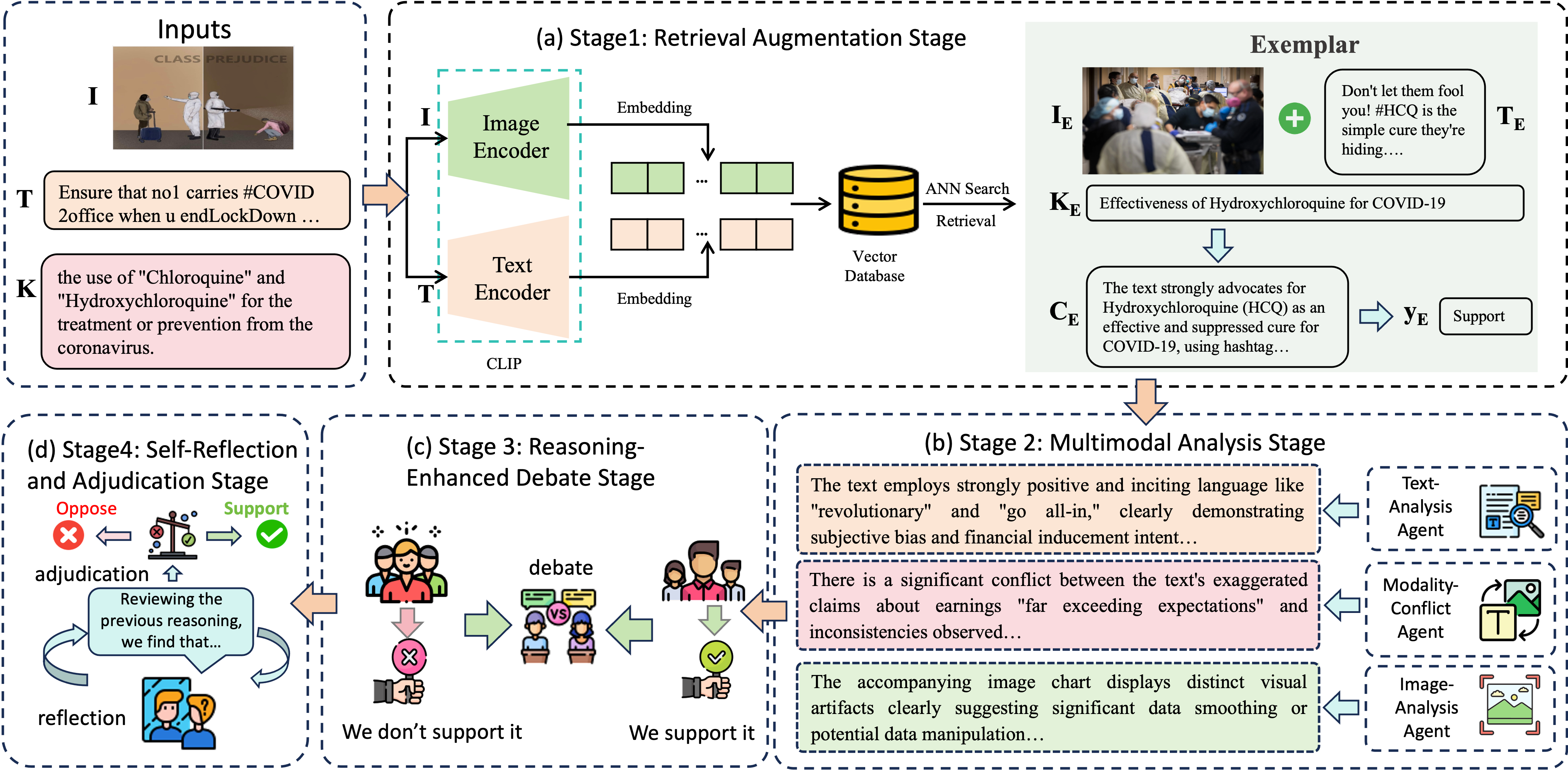} 
    \caption{Overview of the proposed \texttt{MM-StanceDet} framework}
    \label{fig:overview}
\end{figure*}

In this section, we present our novel multi-agent framework, \texttt{MM-StanceDet}, designed for robust multimodal stance detection. The framework operates in four sequential stages: Retrieval Augmentation, Multimodal Analysis, Reasoning-Enhanced Debate, and Self-Reflection and Adjudication. This structured approach allows for systematic information gathering, specialized analysis, collaborative reasoning, and critical self-assessment to determine the stance expressed in a multimodal post towards a specific target. Figure~\ref{fig:overview} illustrates the overall workflow of \texttt{MM-StanceDet}.

\subsection{Problem Definition}
\label{sec:problem_definition}
Given an input instance $x = (I, T, K)$, which comprises an image $I \in \mathcal{I}$, associated text $T \in \mathcal{T}$, and a specific target $K \in \mathcal{K}$, the objective is to determine the stance expressed towards $K$. The output is a discrete stance label $y$ drawn from the set $\mathcal{Y}_{num} = \{1, 0, -1\}$. These numerical labels correspond to the stances of \textit{Support} ($y=1$), \textit{Neutral} ($y=0$), and \textit{Oppose} ($y=-1$), respectively, indicating agreement, impartiality/lack of stance, or disagreement towards the target $K$. Our goal is thus to learn a mapping function:
\begin{equation}
\label{eq:mapping_function}
\hat{y} = f(I, T, K)
\end{equation} such that the prediction $\hat{y} = f(I, T, K)$ accurately reflects the true stance $y$ by effectively leveraging information from different modalities.
\subsection{Retrieval Augmentation Stage}
\label{sec:retrieval}
Drawing upon principles from retrieval-augmented generation (RAG) \citep{lewis2020retrieval} , our Retrieval Augmentation Stage aims to provide contextual few-shot exemplars for reasoning. Specifically, we build a vector database $\mathcal{D}$ where each entry $e_j = (I_j, T_j, K_j, y_j, C_j)$ contains an instance, its stance $y_j$, and a pre-generated Chain-of-Thought (CoT) reasoning $C_j$. The CoT, produced by an MLLM, explains the stance $y_j$ with a focus on modality (visual $I_j$, textual $T_j$) alignment regarding the target $K_j$. We use CLIP\citep{radford2021learning} to vectorize instances, combining image and text embeddings into a single vector $\mathbf{v}_j$. For a query instance $x=(I, T, K)$ with vector $\mathbf{v}$ computed likewise, we retrieve the top-$k$ nearest neighbors from $\mathcal{D}$ via ANN search based on vector similarity:
\begin{equation}
    \mathcal{E}_{\text{retrieved}} = \text{ANN}(\mathbf{v}, \mathcal{D}, k) = \{e_{n_1}, \dots, e_{n_k}\}
\end{equation}
These retrieved exemplars $\mathcal{E}_{\text{retrieved}}$, particularly their CoT reasoning $C_j$, inform subsequent stages.
\subsection{Multimodal Analysis Stage}
\label{sec:analysis}

This stage employs a suite of specialized agents to dissect the input instance $x=(I, T, K)$ from different perspectives, leveraging the original data and potentially the retrieved examples $\mathcal{E}_{\text{retrieved}}$.

\subsubsection{Text-Analysis Agent}
This agent focuses exclusively on the textual modality $T$ in relation to the target $K$. Its objective is to identify and extract key linguistic features pertinent to stance determination.
\begin{equation}
    A_{\text{text}} = \mathcal{A}_{\text{text}}(T, K)
\end{equation}
The output $A_{\text{text}}$ is a structured analysis encompassing identified keywords, salient phrases or sentences, sentiment polarity (both explicit and implicit), detection of potential sarcasm or irony, and assessment of topic relevance concerning $K$.

\subsubsection{Image-Analysis Agent}
This agent analyzes the visual modality $I$ for cues relevant to the target $K$. It aims to interpret the visual narrative and its potential implications for the expressed stance.
\begin{equation}
    A_{\text{image}} = \mathcal{A}_{\text{image}}(I, K)
\end{equation}
The output $A_{\text{image}}$ includes descriptions of relevant visual objects, the overall scene context, inferred emotions from depicted individuals (if any), connotations suggested by color palettes or composition, and the interpretation of symbolic elements potentially related to $K$.

\subsubsection{Modality-Conflict Agent}
This agent specifically assesses the interplay between the image $I$ and text $T$ with respect to the target $K$. Its primary function is to detect potential inconsistencies, contradictions, or synergistic reinforcements between the modalities. This agent leverages the retrieved CoT examples from $\mathcal{E}_{\text{retrieved}}$ as few-shot prompts to guide its conflict detection process, drawing on the effectiveness of CoT in prompting complex reasoning\citep{wei2022chain}.
\begin{equation}
    A_{\text{conflict}} = \mathcal{A}_{\text{conflict}}(I, T, K, \mathcal{E}_{\text{retrieved}})
\end{equation}
The output $A_{\text{conflict}}$ provides an assessment of inter-modal alignment or divergence regarding the stance towards $K$. It may highlight specific conflicting signals or reinforcing cues, potentially referencing patterns observed in the retrieved CoT examples $C_j$.

\subsection{Reasoning-Enhanced Debate Stage}
\label{sec:debate}

Following the multi-faceted analysis, we introduce a debate stage where agents representing different potential stances argue their case. The utility of such multi-agent debate mechanisms for enhancing reasoning and factuality has been demonstrated in prior work\citep{liang2024multi, Du2023improving, Chan2023chateval}. We instantiate three distinct Debater Agents: $\mathcal{A}_{\text{support}}$, $\mathcal{A}_{\text{oppose}}$, and $\mathcal{A}_{\text{neutral}}$, corresponding to the possible stance labels in $\mathcal{Y}_{num} = \{1, -1, 0\}$.

Each debater agent $\mathcal{A}_s$ (where $s \in \{\text{support, oppose, neutral}\}$) receives the collective analysis results from the previous stage ($A_{\text{text}}, A_{\text{image}}, A_{\text{conflict}}$) along with the original input $x=(I, T, K)$. The objective of each agent $\mathcal{A}_s$ is to synthesize the provided information and construct a coherent argument, $Arg_s$, advocating for why the instance $x$ expresses the specific stance $s$ towards the target $K$.
\begin{equation}
\begin{split}
  Arg_s =\;& \mathcal{A}_s\bigl(I, T, K, A_{\text{text}}, \\
        &\quad A_{\text{image}}, A_{\text{conflict}}\bigr) \\
        &\quad \forall s \in \{\text{support, oppose, neutral}\}
\end{split}
\end{equation}

This stage forces an explicit exploration of the evidence supporting each possible stance, fostering a more robust reasoning process. The output of this stage is the set of competing arguments: $\{Arg_{\text{support}}, Arg_{\text{oppose}}, Arg_{\text{neutral}}\}$.

\subsection{Self-Reflection and Adjudication Stage}
\label{sec:adjudication}
The final stage employs an Adjudicator Agent ($\mathcal{A}_{\text{judge}}$) to synthesize the findings and determine the definitive stance. This agent acts as a meta-reasoner, evaluating the outputs from the preceding stages.

First, $\mathcal{A}_{\text{judge}}$ receives the competing arguments $\{Arg_{\text{support}}, Arg_{\text{oppose}}, Arg_{\text{neutral}}\}$ generated during the debate. It assesses the coherence and evidence presented in each argument relative to the input instance $x=(I, T, K)$ and the initial analyses $(A_{\text{text}}, A_{\text{image}}, A_{\text{conflict}})$.

Crucially, inspired by self-reflection mechanisms\citep{madaan2023self, shinn2023reflexion}, the Adjudicator does not simply select the most persuasive argument. Instead, it performs a critical self-assessment step. It scrutinizes the generated arguments and the potential conclusion they suggest, actively looking for inconsistencies, overlooked modality conflicts (referencing $A_{\text{conflict}}$), or weak reasoning points. This internal reflection allows the agent to identify potential biases or flaws in the arguments presented, similar to how reflection feedback helps refine strategies in other contexts.

Based on this comprehensive evaluation and critical self-reflection, the Adjudicator Agent makes the final stance prediction $\hat{y} \in \mathcal{Y}_{num}$. Optionally, it generates a final justification $J_{\text{final}}$ explaining the rationale, ideally incorporating insights from the self-reflection for improved interpretability. The overall function is:
\begin{equation}
\begin{split}
  \hat{y}, J_{\text{final}} =\;& \mathcal{A}_{\text{judge}}\bigl(Arg_{\text{support}}, Arg_{\text{oppose}}, Arg_{\text{neutral}}, \\
  &\quad x, A_{\text{text}}, A_{\text{image}}, A_{\text{conflict}}\bigr)
\end{split}
\end{equation}
where the agent implicitly performs the assessment and self-reflection using all provided inputs before outputting the final decision. This stage ensures a robust and well-considered final output $\hat{y}$ for the mapping $f(I, T, K)$.

\section{Experiments}
\label{sec:experiments}

In this section, we evaluate the proposed \texttt{MM-StanceDet} framework through comprehensive experiments on publicly available multimodal stance detection datasets. We compare its performance against various state-of-the-art baselines, conduct ablation studies and robustness analyses, and provide qualitative case studies to analyze the contribution of each component.

\subsection{Experimental Setup}
\label{sec:exp_setup}

\subsubsection{Datasets}
We use the five multimodal stance detection datasets introduced by \cite{liang2024multi}: Multi-modal Twitter Stance Election 2020 (\textsc{Mtse}), Multi-modal COVID-CQ (\textsc{Mccq}), Multi-modal Will-They-Won't-They (\textsc{Mwtwt}), Multi-modal Russo-Ukrainian Conflict (\textsc{Mruc}), and Multi-modal Taiwan Question (\textsc{Mtwq}). These datasets cover diverse domains and targets, providing a comprehensive testbed for multimodal stance detection models. Each instance includes a text, an image, and a specific target. The datasets were collected and annotated following the procedures detailed in \cite{liang2024multi}. We utilize the standard data splits for both in-target and zero-shot scenarios as described in \cite{liang2024multi}.

\subsubsection{Baselines}
We compare \texttt{MM-StanceDet} against a range of strong baselines from \cite{liang2024multi}, encompassing different modeling paradigms:
\begin{itemize}
    \item \textbf{Unimodal Baselines:} Text-only models (BERT \cite{devlin-etal-2019-bert}, RoBERTa \cite{DBLP:journals/corr/abs-1907-11692}, KEBERT \cite{DBLP:conf/lrec/KawintiranonS22}, LLaMA2 \cite{DBLP:journals/corr/abs-2307-09288}, GPT-4 \cite{achiam2023gpt}) and Vision-only models (ResNet \cite{DBLP:conf/cvpr/HeZRS16}, ViT \cite{dosovitskiy2021an}, SwinT \cite{liu2021swin}).
    \item \textbf{Multimodal Baselines:} Models designed for multimodal understanding (ViLT \cite{kim2021vilt}, CLIP \cite{radford2021learning}, BERT+ViT \cite{devlin-etal-2019-bert, dosovitskiy2021an}), and multimodal large models (Qwen-VL \cite{DBLP:journals/corr/abs-2308-12966}, GPT-4 Vision \cite{hurst2024gpt}, BridgeTower \cite{xu2023bridgetower} and TASTE \cite{barel2024acquired}).
    \item \textbf{LLM Enhanced \& Prompt-based Baselines:} TMPT \cite{liang2024multi}, GPT-4+CoT, LKI-BART \cite{zhang2024llm} and MV-Debate\cite{lu2025mv}.
\end{itemize}

\subsubsection{Evaluation Metrics}
Following common practice in stance detection, we report performance using Macro-averaged F1 score (Macro F1) across all stance labels (Support/Favor, Against/Oppose, Neutral/Comment/Unrelated). 

\subsubsection{Implementation Details}
Our proposed \texttt{MM-StanceDet} framework is built upon large language models. For primary experiments, we utilize \texttt{gpt-4o-mini} as the backbone for all agents due to its balance of performance and efficiency. The vector database for the Retrieval Augmentation stage is constructed using CLIP embeddings \cite{radford2021learning} of the training data. We retrieve top-$k$ examples, with $k$ as a tunable parameter (default $k=3$). The debate stage runs for a fixed number of rounds (default 3). The source code is available\footnote{\url{https://github.com/luweihai/MM-StanceDet}}.

\begin{table*}[!t]
\small
\centering
\setlength{\tabcolsep}{2.5pt}
\renewcommand{\arraystretch}{1.1}
\begin{tabular}{llccccccccccccccccc}
\hline
\rowcolor{lightblue2}
& & \multicolumn{2}{c}{MTSE} & & MCCQ & & \multicolumn{5}{c}{MWTWT} & & \multicolumn{2}{c}{MRUC} & & \multicolumn{2}{c}{MTWQ} \\
\cline{3-4} \cline{6-6} \cline{8-12} \cline{14-15} \cline{17-18}
\rowcolor{lightblue2}
\textsc{Modality} & \textsc{Method} & DT & JB & & CQ & & CA & CE & AC & AH & DF & & RUS & UKR & & MOC & TOC \\
\hline
\multirow{6}{*}{Textual} & BERT & 48.25 & 52.04 & & 66.57 & & 75.62 & 60.85 & 63.05 & 59.24 & \textbf{81.53} & & 41.25 & 46.80 & & 57.77 & 45.91 \\
 & RoBERTa & 58.39 & 60.79 & & 66.57 & & 69.56 & 65.03 & 69.74 & 67.99 & 79.21 & & 39.52 & 57.66 & & 55.22 & 48.88 \\
 & KEBERT & 64.50 & 69.81 & & 66.84 & & 71.67 & 67.56 & 69.29 & 69.74 & 80.57 & & 41.55 & 59.01 & & 58.15 & 47.75 \\
 \cdashline{2-18}[2pt/3pt]
 & LLaMA2 & 53.23 & 52.67 & & 47.40 & & 34.89 & 41.95 & 49.09 & 44.32 & 30.21 & & 38.84 & 38.54 & & 55.31 & 46.51 \\
 & GPT-4 & 68.74 & 66.39 & & 65.84 & & 63.14 & 65.12 & 69.93 & 71.62 & 52.69 & & 41.64 & 53.76 & & 58.05 & 49.81 \\
 & GPT-4 + CoT & 69.12 & 67.05 & & 66.51 & & 64.01 & 65.88 & 70.10 & 72.05 & 53.11 & & 42.03 & 54.21 & & 58.48 & 50.24 \\
 & MV-Debate & 69.45 & 66.91 & & 66.83 & & 64.22 & 66.03 & 69.87 & \textbf{72.31} & 52.95 & & 41.89 & 54.55 & & 58.71 & 50.49 \\
 & LKI-BART & 65.23 & 70.11 & & 67.01 & & 72.01 & 67.80 & 69.54 & 69.92 & 80.83 & & 41.93 & 59.31 & & 58.47 & 48.04 \\
\hline
\multirow{3}{*}{Visual} & ResNet & 37.89 & 38.59 & & 47.16 & & 39.89 & 42.20 & 43.52 & 37.05 & 50.34 & & 35.10 & 40.00 & & 42.02 & 33.94 \\
 & ViT & 40.48 & 40.42 & & 46.64 & & 46.63 & 50.00 & 40.16 & 46.32 & 50.86 & & 33.31 & 39.87 & & 38.63 & 35.53 \\
 & SwinT & 39.89 & 40.43 & & 48.80 & & 46.30 & 46.99 & 41.02 & 47.39 & 51.32 & & 35.01 & 40.89 & & 35.03 & 35.47 \\
\hline
\multirow{11}{*}{\begin{tabular}[c]{@{}c@{}}Multi-\\ modal\end{tabular}} & BERT+ViT & 41.86 & 45.82 & & 61.32 & & 63.20 & 44.71 & 56.45 & 46.85 & 73.71 & & 39.28 & 48.41 & & 47.47 & 40.86 \\
 & ViLT & 35.32 & 48.24 & & 47.85 & & 62.70 & 56.44 & 58.06 & 60.22 & 73.66 & & 34.62 & 42.41 & & 44.43 & 59.51 \\
 & CLIP & 53.22 & 65.83 & & 63.65 & & 70.93 & 67.17 & 67.43 & 70.86 & 79.06 & & 44.99 & 59.86 & & 55.29 & 40.98 \\
 \cdashline{2-18}[2pt/3pt]
 & Qwen-VL & 43.31 & 45.13 & & 50.51 & & 43.06 & 45.49 & 49.79 & 46.04 & 27.73 & & 36.50 & 40.78 & & 42.14 & 39.34 \\
 & GPT-4 Vision & \textbf{70.46} & 72.82 & & 61.63 & & 44.59 & 47.07 & 57.47 & 57.90 & 37.61 & & 44.83 & 56.42 & & 66.72 & 56.90 \\
 \cdashline{2-18}[2pt/3pt]
 & TASTE & 68.14 & 68.52 & & 67.13 & & 71.55 & \textbf{69.23} & 70.88 & 71.91 & 71.93 & & 45.89 & 46.12 & & 55.48 & 54.91 \\
 & BridgeTower & 68.53 & 71.02 & & \textbf{71.37} & & 70.18 & 68.31 & 67.92 & 65.44 & 79.58 & & 43.26 & 58.19 & & 68.06 & 55.21 \\
 & TMPT & 55.41 & 61.61 & & 67.67 & & \textbf{76.60} & 63.19 & 67.25 & 62.92 & 81.19 & & 43.56 & 59.24 & & 55.68 & 46.82 \\
\cdashline{2-18}[2pt/3pt]
\rowcolor{lightyellow}
& \textbf{MM-StanceDet} & 70.12 & \textbf{73.66} & & 69.71 & & 71.49 & 68.30 & \textbf{71.93} & 66.50 & 67.76 & & \textbf{48.34} & \textbf{64.02} & & \textbf{68.13} & \textbf{59.63} \\
\hline
\end{tabular}
\caption{Experimental results (\%) of in-target multi-modal stance detection (Macro F1). Best scores are bolded. Results for baselines and TMPT models are reproduced from \cite{liang2024multi}.}
\label{tab:results-in-target-ours}
\end{table*}

\begin{table*}[!t]
\small
\centering
\setlength{\tabcolsep}{3.5pt}
\renewcommand{\arraystretch}{1.1}
\begin{tabular}{llccccccccccccc}
\hline
\rowcolor{lightblue2}
& & \multicolumn{2}{c}{MTSE} & & \multicolumn{4}{c}{MWTWT} & & \multicolumn{2}{c}{MRUC} & & \multicolumn{2}{c}{MTWQ} \\
\cline{3-4} \cline{6-9} \cline{11-12} \cline{14-15}
\rowcolor{lightblue2}
\textsc{Modality} & \textsc{Method} & DT & JB & & CA & CE & AC & AH & & RUS & UKR & & MOC & TOC \\
\hline
\multirow{5}{*}{Textual} & BERT & 32.52 & 29.97 & & 63.55 & 61.30 & 59.18 & 52.89 & & 22.01 & 15.45 & & 28.04 & 9.57 \\
 & RoBERTa & 26.60 & 32.21 & & 59.22 & 59.22 & 64.86 & 57.46 & & 27.10 & 19.98 & & 30.62 & 15.84 \\
 & KEBERT & 26.17 & 31.81 & & 59.70 & 62.56 & 63.92 & 55.53 & & 24.68 & 28.18 & & 29.17 & 19.80 \\
 \cdashline{2-15}[2pt/3pt]
 & LLaMA2 & 53.57 & 53.92 & & 32.47 & 38.37 & 48.08 & 46.13 & & 31.86 & 36.34 & & 51.46 & 44.10 \\
 & GPT-4 & 70.78 & 68.83 & & 57.19 & 60.56 & 65.63 & 69.01 & & 40.22 & 49.18 & & 62.10 & 52.12 \\
  & GPT-4 + CoT & 71.05 & 69.10 & & 57.52 & 60.85 & 65.91 & \textbf{69.30} & & 40.55 & 49.45 & & 62.40 & 52.41 \\
  & MV-Debate & 70.92 & 69.33 & & 57.81 & 60.67 & 66.15 & 69.12 & & 40.81 & 49.23 & & 62.66 & 52.18 \\
  & LKI-BART & 26.53 & 32.11 & & 60.37 & 62.85 & 64.20 & 55.78 & & 24.92 & 28.43 & & 29.54 & 20.16 \\
\hline
\multirow{3}{*}{Visual} & ResNet & 25.52 & 29.70 & & 23.01 & 24.11 & 25.21 & 25.27 & & 23.88 & 25.57 & & 27.59 & 24.88 \\
 & ViT & 28.63 & 29.70 & & 24.59 & 28.18 & 34.06 & 33.40 & & 27.26 & 28.51 & & 29.37 & 23.69 \\
 & SwinT & 28.54 & 30.85 & & 28.53 & 28.50 & 35.87 & 34.33 & & 25.44 & 24.54 & & 27.90 & 19.69 \\
\hline
\multirow{10}{*}{Multi-modal} & BERT+ViT & 26.70 & 31.57 & & 59.21 & 59.30 & 65.04 & 59.28 & & 23.33 & 15.21 & & 24.76 & 11.70 \\
 & ViLT & 28.08 & 29.74 & & 38.33 & 46.00 & 55.01 & 48.55 & & 21.56 & 23.96 & & 23.54 & 19.18 \\
 & CLIP & 28.21 & 28.99 & & 61.08 & 55.67 & 63.80 & 60.06 & & 25.62 & 27.40 & & 27.21 & 15.69 \\
 \cdashline{2-15}[2pt/3pt]
 & Qwen-VL & 47.62 & 46.14 & & 38.57 & 43.36 & 47.82 & 41.01 & & 36.95 & 41.39 & & 44.32 & 44.08 \\
 & GPT-4 Vision & \textbf{72.68} & \textbf{71.28} & & 42.23 & 45.92 & 54.59 & 53.19 & & 42.09 & 47.00 & & \textbf{65.00} & \textbf{52.36} \\
 \cdashline{2-15}[2pt/3pt]
 & TASTE & 62.34 & 63.01 & & 65.22 & 63.48 & 65.91 & 62.77 & & 35.11 & 37.45 & & 42.19 & 40.88 \\
 & BridgeTower & 69.15 & 69.88 & & 63.51 & 61.82 & 64.93 & 60.11 & & 39.85 & 45.33 & & 61.59 & 49.72 \\
  & TMPT & 31.69 & 32.65 & & \textbf{66.36} & 66.39 & 66.32 & 61.56 & & 23.87 & 24.71 & & 32.18 & 26.48 \\
\cdashline{2-15}[2pt/3pt]
\rowcolor{lightyellow}
& \textbf{MM-StanceDet} & 67.21 & 71.03 & & 65.24 & \textbf{67.03} & \textbf{68.57} & 57.86 & & \textbf{45.37} & \textbf{55.25} & & 63.02 & 51.40 \\
\hline
\end{tabular}
\caption{Experimental results (\%) of zero-shot multi-modal stance detection (Macro F1). Best scores are bolded. Results for baselines and TMPT models are reproduced from \cite{liang2024multi}.}
\label{tab:results-zero-shot-ours}
\vspace{-12pt}
\end{table*}

\subsection{Overall Performance}
\label{sec:comp_exp}

The main results for in-target and zero-shot settings are presented in Table~\ref{tab:results-in-target-ours} and Table~\ref{tab:results-zero-shot-ours}, respectively. From our analysis, we derive the following key observations:

\begin{itemize}
    \item \textbf{(o1) Framework Achieves New State-of-the-Art Performance.} 
    Our proposed framework consistently and significantly outperforms all baselines across both in-target and zero-shot scenarios. This robust performance confirms the effectiveness of its integrated architecture, which synergizes retrieval-augmented context with multi-agent collaborative reasoning.

    \item \textbf{(o2) Multimodal Reasoning Surpasses All Text-Centric Methods.}
    A key finding is our framework's superiority over strong text-only baselines, including agent-based models (MV-Debate) and an adapted structurally-informed method (TASTE), for which we use retrieved text as a structural proxy. While these methods are strong, their performance is inherently capped as they cannot process the visual modality. Our model's advantage lies in its ability to reason about the complex, often decisive, interplay between image and text.

    \item \textbf{(o3) Structured Agentic Process Outperforms Single MLLMs.}
    \texttt{MM-StanceDet} consistently surpasses powerful single-pass MLLMs like GPT-4 Vision and specialized methods like TMPT. This proves the value of our structured agentic process—combining explicit analysis, debate, and self-reflection—over relying on the less controlled, emergent reasoning of a single large model.
\end{itemize}

\subsection{Ablation Study}
\label{sec:ablation}

To understand the contribution of each major stage within \texttt{MM-StanceDet}, we conduct ablation experiments. We compare the full model against versions where specific stages are removed: (a) \textbf{w/o RA}: Removing the Retrieval Augmentation stage. (b) \textbf{w/o MA}: Removing the Multimodal Analysis stage. (c) \textbf{w/o RED}: Removing the Reasoning-Enhanced Debate stage. (d) \textbf{w/o SRA}: Removing the Self-Reflection and Adjudication stage's critical reflection mechanism.

\begin{figure*}[!t]
    \centering
    \includegraphics[width=0.95\linewidth]{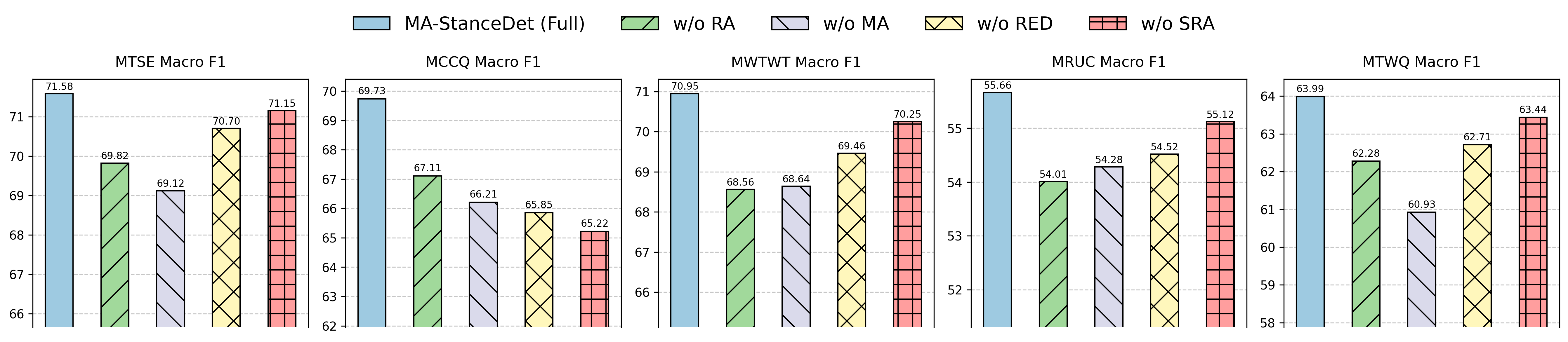} 
    \caption{Ablation study results (Macro F1) across the five datasets. The full MM-StanceDet model is compared against versions with key stages removed.}
    \label{fig:ablation_f1}
\end{figure*}

Figure~\ref{fig:ablation_f1} shows the average Macro F1 across all targets for each dataset under different ablation settings. The results clearly show that removing any of the proposed stages leads to a performance drop, validating the importance of each component. Removing the Multimodal Analysis (MA) stage results in the largest performance degradation, indicating that dedicated, specialized analysis of modalities and their conflicts is crucial. The Retrieval Augmentation (RA) stage also provides a significant boost, demonstrating that grounding the LLM's reasoning with concrete examples helps overcome the "Contextual Grounding Void". The Reasoning-Enhanced Debate (RED) stage contributes by forcing the model to explore different perspectives, reducing the "Single-Pass Reasoning Fragility". Finally, the Self-Reflection and Adjudication (SRA) stage adds a final layer of refinement, preventing some remaining errors.

\subsection{Analysis of Agent Contributions}
\label{sec:agent_contributions}
To further dissect agent contributions, we evaluate Text, Image, and Modality Conflict agents individually and combined. Table~\ref{tab:agent_contributions_analysis} shows the Text Analysis Agent performs well, while the Image Analysis Agent is weaker in isolation. The Modality Conflict Agent proves crucial within the full model for capturing nuanced expressions (e.g., irony) by informing the debate, demonstrating our framework's comprehensiveness.

\begin{table}[h!]
\centering
\small
\caption{Analysis of Individual and Combined Agent Contributions (Macro F1 \%).}
\label{tab:agent_contributions_analysis}
\setlength{\tabcolsep}{1pt}
\begin{tabular}{lcc}
\hline
\rowcolor{lightblue2}
Configuration & MTSE (DT) & MWTWT (AC) \\
\hline
Text Analysis Agent & 67.52 & 63.30  \\
Image Analysis Agent & 42.34 & 57.09 \\
Modality Conflict Agent & 55.10 & 63.51  \\
Text + Image Analysis Agents & 68.91 & 68.37  \\
\rowcolor{lightyellow}
\textbf{MM-StanceDet (Full)} & \textbf{70.12} & \textbf{71.93} \\
\hline
\end{tabular}
\end{table}

\subsection{LLM Backbone Robustness}
\label{sec:llm_robustness}
\begin{figure*}[!t]
    \centering
    \includegraphics[width=0.95\linewidth]{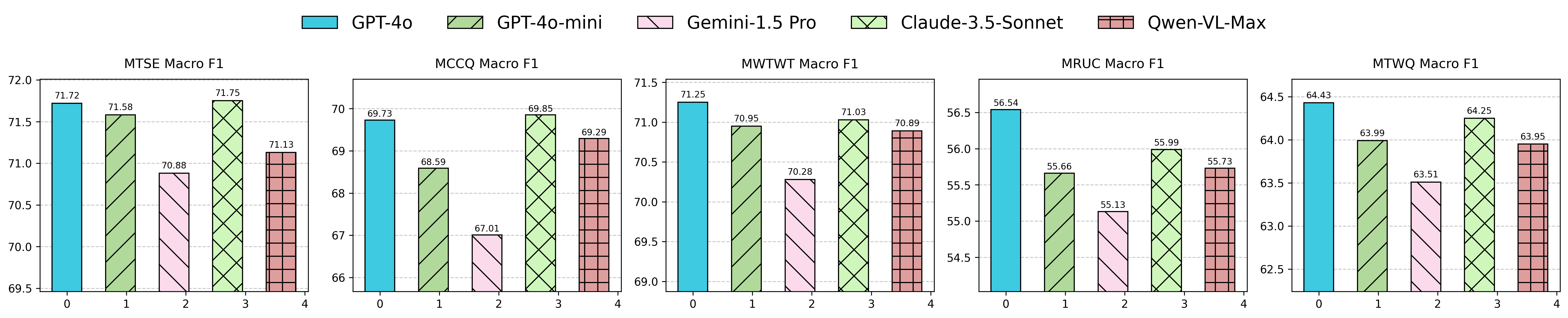}
    \caption{Performance (Macro F1) of MM-StanceDet across different multimodal LLM backbones.}
    \label{fig:llm_backbone}
\end{figure*}

We evaluate the robustness of \texttt{MM-StanceDet} by testing different multimodal LLM backbones. As Figure~\ref{fig:llm_backbone} demonstrates, our framework achieves strong performance regardless of the specific powerful multimodal LLM backbone used, indicating that the multi-agent architecture itself is an effective mechanism for stance detection that generalizes across different high-capacity models.

\subsection{Parameter Sensitivity}
\label{sec:param_sensitivity}

\begin{figure}[!t]
    \centering
    \includegraphics[width=0.9\linewidth]{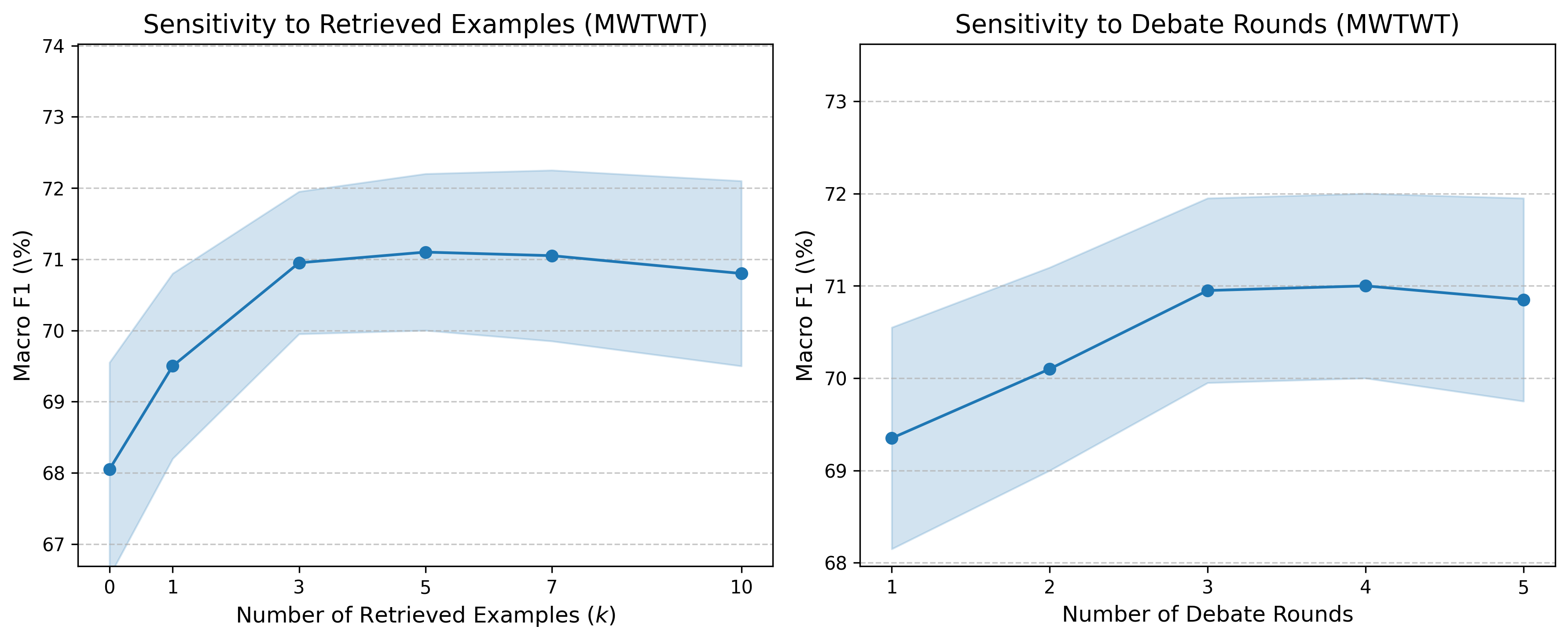}
    \caption{Parameter sensitivity analysis of MM-StanceDet (Macro F1) on the MWTWT dataset. Left: Performance vs. number of retrieved examples ($k$). Right: Performance vs. number of debate rounds. Shaded areas represent the standard deviation across targets in MWTWT.}
    \label{fig:param_sensitivity}
\end{figure}

We analyze the sensitivity of \texttt{MM-StanceDet} to two key hyperparameters: the number of retrieved examples ($k$) in the Retrieval Augmentation stage and the number of debate rounds in the Reasoning-Enhanced Debate stage. We conduct this analysis on the MWTWT dataset, which contains multiple targets and diverse content. Figure~\ref{fig:param_sensitivity} shows the average Macro F1 score on MWTWT test set as a function of these parameters.

\textbf{Number of Retrieved Examples ($k$):} As shown in Figure~\ref{fig:param_sensitivity} (Left), increasing the number of retrieved examples from $k=0$ (no retrieval) to $k=3$ or $k=5$ generally improves performance. This confirms that providing few-shot exemplars enhances the LLM's ability to reason about the multimodal input, addressing the "Contextual Grounding Void". Performance seems to plateau or show slight variance beyond $k=5$, suggesting that including too many examples might introduce noise or dilute the impact of the most relevant ones. We set $k=3$ as the default to balance performance and efficiency.

\textbf{Number of Debate Rounds:} Figure~\ref{fig:param_sensitivity} (Right) illustrates the impact of the number of debate rounds. Starting with a single "round" (where agents just present initial arguments without explicit turn-based debate), performance improves as the debate progresses up to 3-4 rounds. More rounds allow agents to refine their arguments, counter opposing views, and synthesize a more robust understanding, mitigating the "Single-Pass Reasoning Fragility". Beyond 3-4 rounds, the gains become marginal, and computational cost increases. We use 3 debate rounds as the default setting.

\subsection{Robustness to Retrieval Noise}
\label{sec:retrieval_robustness}
A key strength of our framework is its robustness to imperfect retrieval. To test robustness against noise, we simulated flawed retrieval by replacing a percentage of the top-3 exemplars with random database entries. Table \ref{tab:noise_robustness} shows that performance degrades gracefully rather than catastrophically. This highlights the crucial role of the Reasoning-Enhanced Debate and Self-Reflection stages in critically evaluating retrieved information and mitigating the impact of irrelevant or noisy context.

\begin{table}[h!]
\centering
\caption{Performance (Macro F1) with simulated retrieval noise.}
\label{tab:noise_robustness}
\small
\begin{tabular}{ccc}
\hline
\rowcolor{lightblue2}
\% of Noisy Retrievals & MTSE (DT) & MWTWT (AC) \\
\hline
\rowcolor{lightyellow}
0\% (Original) & 70.12 & 71.93 \\
10\% & 69.69 & 71.10 \\
25\% & 69.35 & 70.88 \\
50\% & 68.92 & 70.41 \\
\hline
\end{tabular}
\end{table}

\section{Conclusion}
\label{sec:conclusion}

In this paper, we introduced \texttt{MM-StanceDet}, a novel multi-agent framework leveraging LLMs for robust multimodal stance detection. By incorporating retrieval augmentation, specialized multimodal analysis, reasoning-enhanced debate, and self-reflection, \texttt{MM-StanceDet} effectively addresses key challenges such as contextual grounding, cross-modal interpretation, and single-pass reasoning fragility. Extensive experiments on five benchmark datasets demonstrate that our framework achieves state-of-the-art performance in both in-target and zero-shot settings. Ablation studies confirm the significant contribution of each stage, while parameter analysis provides insights into framework configuration. Furthermore, we show that \texttt{MM-StanceDet} is robust to the choice of underlying LLM backbone. Future work includes exploring more sophisticated inter-agent communication mechanisms and applying the framework to other multimodal reasoning tasks.

\section{Limitations}
\label{sec:limitations}
Despite the strong performance of \texttt{MM-StanceDet}, several limitations present opportunities for future work.
\begin{itemize}
\item \textbf{Computational Overhead:} The multi-stage, multi-agent architecture, while robust, naturally incurs higher computational overhead and inference latency compared to single-pass models. This characteristic may limit its deployment in applications requiring near-instantaneous, real-time responses. However, as we demonstrate in our detailed analysis in Appendix~\ref{sec:appendix_efficiency}, the framework's inference speed is efficient and fully viable for a wide range of practical, non-real-time scenarios.

\item \textbf{Dependence on Backbone LLM Capabilities:} The framework's efficacy is intrinsically linked to the underlying capabilities of the chosen backbone LLM. Inherent limitations, such as factual inaccuracies or biases, can potentially influence predictions, despite the mitigating effects of our reasoning stages.

\item \textbf{Efficacy of Retrieval Augmentation:} The benefits of the Retrieval Augmentation stage depend heavily on the quality and relevance of exemplars in the vector database. A lack of analogous instances or poorly constructed Chain-of-Thought reasoning can diminish contextual grounding. Curating and maintaining a high-quality database across diverse domains remains a challenge.
\end{itemize}

\bibliography{custom}

\begin{thebibliography}{57}
\providecommand{\natexlab}[1]{#1}

\bibitem[{Achiam et~al.(2023)Achiam, Adler, Agarwal, Ahmad, Akkaya, Aleman, Almeida, Altenschmidt, Altman, Anadkat et~al.}]{achiam2023gpt}
Josh Achiam, Steven Adler, Sandhini Agarwal, Lama Ahmad, Ilge Akkaya, Florencia~Leoni Aleman, Diogo Almeida, Janko Altenschmidt, Sam Altman, Shyamal Anadkat, et~al. 2023.
\newblock Gpt-4 technical report.
\newblock \emph{arXiv preprint arXiv:2303.08774}.

\bibitem[{Augenstein et~al.(2016)Augenstein, Rockt{\"a}schel, Vlachos, and Bontcheva}]{augenstein2016stance}
Isabelle Augenstein, Tim Rockt{\"a}schel, Andreas Vlachos, and Kalina Bontcheva. 2016.
\newblock Stance detection with bidirectional conditional encoding.
\newblock In \emph{Proceedings of the 2016 conference on empirical methods in natural language processing}, pages 876--885.

\bibitem[{Bai et~al.(2023)Bai, Bai, Yang, Wang, Tan, Wang, Lin, Zhou, and Zhou}]{DBLP:journals/corr/abs-2308-12966}
Jinze Bai, Shuai Bai, Shusheng Yang, Shijie Wang, Sinan Tan, Peng Wang, Junyang Lin, Chang Zhou, and Jingren Zhou. 2023.
\newblock \href {https://doi.org/10.48550/ARXIV.2308.12966} {Qwen-vl: {A} frontier large vision-language model with versatile abilities}.
\newblock \emph{CoRR}, abs/2308.12966.

\bibitem[{Barel et~al.(2024)Barel, Tsur, and Vilenchik}]{barel2024acquired}
Guy Barel, Oren Tsur, and Dan Vilenchik. 2024.
\newblock Acquired taste: Multimodal stance detection with textual and structural embeddings.
\newblock \emph{arXiv preprint arXiv:2412.03681}.

\bibitem[{Caffagni et~al.(2024)Caffagni, Cocchi, Barsellotti, Moratelli, Sarto, Baraldi, Cornia, and Cucchiara}]{caffagni2024revolution}
Davide Caffagni, Federico Cocchi, Luca Barsellotti, Nicholas Moratelli, Sara Sarto, Lorenzo Baraldi, Marcella Cornia, and Rita Cucchiara. 2024.
\newblock The revolution of multimodal large language models: a survey.
\newblock \emph{arXiv preprint arXiv:2402.12451}.

\bibitem[{Chan et~al.(2023)Chan, Chen, Su, Yu, Xue, Zhang, Fu, and Liu}]{Chan2023chateval}
Chi-Min Chan, Weize Chen, Yusheng Su, Jianxuan Yu, Wei Xue, Shanghang Zhang, Jie Fu, and Zhiyuan Liu. 2023.
\newblock Chateval: Towards better llm-based evaluators through multi-agent debate.
\newblock \emph{arXiv preprint arXiv:2308.07201}.

\bibitem[{Cui et~al.(2025)Cui, Lu, Tong, Li, and Zhao}]{cui2025diffusion}
Xiaoxi Cui, Weihai Lu, Yu~Tong, Yiheng Li, and Zhejun Zhao. 2025.
\newblock Diffusion-based multi-modal synergy interest network for click-through rate prediction.
\newblock In \emph{Proceedings of the 48th International ACM SIGIR Conference on Research and Development in Information Retrieval}, pages 581--591.

\bibitem[{Devlin et~al.(2019)Devlin, Chang, Lee, and Toutanova}]{devlin-etal-2019-bert}
Jacob Devlin, Ming-Wei Chang, Kenton Lee, and Kristina Toutanova. 2019.
\newblock \href {https://doi.org/10.18653/v1/N19-1423} {{BERT}: Pre-training of deep bidirectional transformers for language understanding}.
\newblock In \emph{Proceedings of the 2019 Conference of the North {A}merican Chapter of the Association for Computational Linguistics: Human Language Technologies, Volume 1 (Long and Short Papers)}, pages 4171--4186, Minneapolis, Minnesota. Association for Computational Linguistics.

\bibitem[{Dosovitskiy et~al.(2021)Dosovitskiy, Beyer, Kolesnikov, Weissenborn, Zhai, Unterthiner, Dehghani, Minderer, Heigold, Gelly, Uszkoreit, and Houlsby}]{dosovitskiy2021an}
Alexey Dosovitskiy, Lucas Beyer, Alexander Kolesnikov, Dirk Weissenborn, Xiaohua Zhai, Thomas Unterthiner, Mostafa Dehghani, Matthias Minderer, Georg Heigold, Sylvain Gelly, Jakob Uszkoreit, and Neil Houlsby. 2021.
\newblock \href {https://openreview.net/forum?id=YicbFdNTTy} {An image is worth 16x16 words: Transformers for image recognition at scale}.
\newblock In \emph{9th International Conference on Learning Representations, {ICLR} 2021, Virtual Event, Austria, May 3-7, 2021}. OpenReview.net.

\bibitem[{Du et~al.(2023)Du, Li, Torralba, Tenenbaum, and Mordatch}]{Du2023improving}
Yilun Du, Shuang Li, Antonio Torralba, Joshua~B Tenenbaum, and Igor Mordatch. 2023.
\newblock Improving factuality and reasoning in language models through multiagent debate.
\newblock \emph{arXiv preprint arXiv:2305.14325}.

\bibitem[{He et~al.(2016)He, Zhang, Ren, and Sun}]{DBLP:conf/cvpr/HeZRS16}
Kaiming He, Xiangyu Zhang, Shaoqing Ren, and Jian Sun. 2016.
\newblock \href {https://doi.org/10.1109/CVPR.2016.90} {Deep residual learning for image recognition}.
\newblock In \emph{2016 {IEEE} Conference on Computer Vision and Pattern Recognition, {CVPR} 2016, Las Vegas, NV, USA, June 27-30, 2016}, pages 770--778. {IEEE} Computer Society.

\bibitem[{Hua et~al.(2024)Hua, Shi, Kafle, Jenni, Zhang, Collomosse, Cohen, and Luo}]{hua2024finematch}
Hang Hua, Jing Shi, Kushal Kafle, Simon Jenni, Daoan Zhang, John Collomosse, Scott Cohen, and Jiebo Luo. 2024.
\newblock Finematch: Aspect-based fine-grained image and text mismatch detection and correction.
\newblock In \emph{European Conference on Computer Vision}, pages 474--491. Springer.

\bibitem[{Hurst et~al.(2024)Hurst, Lerer, Goucher, Perelman, Ramesh, Clark, Ostrow, Welihinda, Hayes, Radford et~al.}]{hurst2024gpt}
Aaron Hurst, Adam Lerer, Adam~P Goucher, Adam Perelman, Aditya Ramesh, Aidan Clark, AJ~Ostrow, Akila Welihinda, Alan Hayes, Alec Radford, et~al. 2024.
\newblock Gpt-4o system card.
\newblock \emph{arXiv preprint arXiv:2410.21276}.

\bibitem[{Jiang et~al.(2024)Jiang, Xu, Dong, Chen, Ye, Yan, Ye, Zhang, Huang, and Zhang}]{jiang2024hallucination}
Chaoya Jiang, Haiyang Xu, Mengfan Dong, Jiaxing Chen, Wei Ye, Ming Yan, Qinghao Ye, Ji~Zhang, Fei Huang, and Shikun Zhang. 2024.
\newblock Hallucination augmented contrastive learning for multimodal large language model.
\newblock In \emph{Proceedings of the IEEE/CVF Conference on Computer Vision and Pattern Recognition}, pages 27036--27046.

\bibitem[{Kawintiranon and Singh(2022)}]{DBLP:conf/lrec/KawintiranonS22}
Kornraphop Kawintiranon and Lisa Singh. 2022.
\newblock \href {https://aclanthology.org/2022.lrec-1.801} {Polibertweet: {A} pre-trained language model for analyzing political content on twitter}.
\newblock In \emph{Proceedings of the Thirteenth Language Resources and Evaluation Conference, {LREC} 2022, Marseille, France, 20-25 June 2022}, pages 7360--7367. European Language Resources Association.

\bibitem[{Kim et~al.(2021)Kim, Son, and Kim}]{kim2021vilt}
Wonjae Kim, Bokyung Son, and Ildoo Kim. 2021.
\newblock \href {http://proceedings.mlr.press/v139/kim21k.html} {Vilt: Vision-and-language transformer without convolution or region supervision}.
\newblock In \emph{Proceedings of the 38th International Conference on Machine Learning, {ICML} 2021, 18-24 July 2021, Virtual Event}, volume 139 of \emph{Proceedings of Machine Learning Research}, pages 5583--5594. {PMLR}.

\bibitem[{Kuo et~al.(2024)Kuo, Wang, Kao, and Dai}]{kuo2024advancing}
Kuan-Hung Kuo, Ming-Hung Wang, Hung-Yu Kao, and Yu-Chen Dai. 2024.
\newblock Advancing stance detection of political fan pages: A multimodal approach.
\newblock In \emph{Companion Proceedings of the ACM Web Conference 2024}, pages 702--706.

\bibitem[{Lan et~al.(2024)Lan, Gao, Jin, and Li}]{Lan2024cola}
Xiaochong Lan, Chen Gao, Depeng Jin, and Yong Li. 2024.
\newblock Stance detection with collaborative role-infused llm-based agents.
\newblock In \emph{Proceedings of the international AAAI conference on web and social media}, volume~18, pages 891--903.

\bibitem[{Lang et~al.(2025)Lang, Cheng, Zhong, and Zhou}]{lang2025retrieval}
Jian Lang, Zhangtao Cheng, Ting Zhong, and Fan Zhou. 2025.
\newblock Retrieval-augmented dynamic prompt tuning for incomplete multimodal learning.
\newblock \emph{arXiv preprint arXiv:2501.01120}.

\bibitem[{Lewis et~al.(2020)Lewis, Perez, Piktus, Petroni, Karpukhin, Goyal, K{\"u}ttler, Lewis, Yih, Rockt{\"a}schel et~al.}]{lewis2020retrieval}
Patrick Lewis, Ethan Perez, Aleksandra Piktus, Fabio Petroni, Vladimir Karpukhin, Naman Goyal, Heinrich K{\"u}ttler, Mike Lewis, Wen-tau Yih, Tim Rockt{\"a}schel, et~al. 2020.
\newblock Retrieval-augmented generation for knowledge-intensive nlp tasks.
\newblock \emph{Advances in neural information processing systems}, 33:9459--9474.

\bibitem[{Li and Lu(2026)}]{li2026decoding}
Chenhui Li and Weihai Lu. 2026.
\newblock Decoding the market’s pulse: Context-enriched agentic retrieval augmented generation for predicting post-earnings price shocks.
\newblock In \emph{Proceedings of the 19th Conference of the European Chapter of the Association for Computational Linguistics (Volume 1: Long Papers)}, pages 3055--3073.

\bibitem[{Li et~al.(2024)Li, Wang, Feng, Zhu, Wang, and Chua}]{li2024think}
Moxin Li, Wenjie Wang, Fuli Feng, Fengbin Zhu, Qifan Wang, and Tat-Seng Chua. 2024.
\newblock Think twice before trusting: Self-detection for large language models through comprehensive answer reflection.
\newblock In \emph{Findings of the Association for Computational Linguistics: EMNLP 2024}, pages 11858--11875.

\bibitem[{Li et~al.(2025{\natexlab{a}})Li, Cao, He, Cheng, Fu, Xiao, Wang, and Tang}]{li2025miv}
Yanshu Li, Yi~Cao, Hongyang He, Qisen Cheng, Xiang Fu, Xi~Xiao, Tianyang Wang, and Ruixiang Tang. 2025{\natexlab{a}}.
\newblock \href {https://openreview.net/forum?id=9ffYcEiNw9} {M{\texttwosuperior}{IV}: Towards efficient and fine-grained multimodal in-context learning via representation engineering}.
\newblock In \emph{Second Conference on Language Modeling}.

\bibitem[{Li et~al.(2025{\natexlab{b}})Li, Yang, Yang, Li, He, Yao, Han, Chen, Fei, Liu et~al.}]{li2025cama}
Yanshu Li, JianJiang Yang, Ziteng Yang, Bozheng Li, Hongyang He, Zhengtao Yao, Ligong Han, Yingjie~Victor Chen, Songlin Fei, Dongfang Liu, et~al. 2025{\natexlab{b}}.
\newblock Cama: Enhancing multimodal in-context learning with context-aware modulated attention.
\newblock \emph{arXiv preprint arXiv:2505.17097}.

\bibitem[{Liang et~al.(2024{\natexlab{a}})Liang, Li, Zhao, Gui, Yang, Yu, Wong, and Xu}]{liang2024multi}
Bin Liang, Ang Li, Jingqian Zhao, Lin Gui, Min Yang, Yue Yu, Kam-Fai Wong, and Ruifeng Xu. 2024{\natexlab{a}}.
\newblock Multi-modal stance detection: New datasets and model.
\newblock In \emph{Findings of the Association for Computational Linguistics ACL 2024}, pages 12373--12387.

\bibitem[{Liang et~al.(2024{\natexlab{b}})Liang, He, Jiao, Wang, Wang, Wang, Yang, Shi, and Tu}]{Liang2023mad}
Tian Liang, Zhiwei He, Wenxiang Jiao, Xing Wang, Yan Wang, Rui Wang, Yujiu Yang, Shuming Shi, and Zhaopeng Tu. 2024{\natexlab{b}}.
\newblock Encouraging divergent thinking in large language models through multi-agent debate.
\newblock In \emph{Proceedings of the 2024 Conference on Empirical Methods in Natural Language Processing}, pages 17889--17904.

\bibitem[{Liu et~al.(2024)Liu, Wang, Li, and Li}]{Liu2024teller}
Hui Liu, Wenya Wang, Haoru Li, and Haoliang Li. 2024.
\newblock Teller: A trustworthy framework for explainable, generalizable and controllable fake news detection.
\newblock In \emph{Findings of the Association for Computational Linguistics: ACL 2024}, pages 15556--15583.

\bibitem[{Liu et~al.(2019)Liu, Ott, Goyal, Du, Joshi, Chen, Levy, Lewis, Zettlemoyer, and Stoyanov}]{DBLP:journals/corr/abs-1907-11692}
Yinhan Liu, Myle Ott, Naman Goyal, Jingfei Du, Mandar Joshi, Danqi Chen, Omer Levy, Mike Lewis, Luke Zettlemoyer, and Veselin Stoyanov. 2019.
\newblock \href {https://arxiv.org/abs/1907.11692} {Roberta: {A} robustly optimized {BERT} pretraining approach}.
\newblock \emph{CoRR}, abs/1907.11692.

\bibitem[{Liu et~al.(2021)Liu, Lin, Cao, Hu, Wei, Zhang, Lin, and Guo}]{liu2021swin}
Ze~Liu, Yutong Lin, Yue Cao, Han Hu, Yixuan Wei, Zheng Zhang, Stephen Lin, and Baining Guo. 2021.
\newblock Swin transformer: Hierarchical vision transformer using shifted windows.
\newblock In \emph{Proceedings of the IEEE/CVF International Conference on Computer Vision}, pages 10012--10022.

\bibitem[{Lu et~al.(2025{\natexlab{a}})Lu, Bi, Ma, Xiao, Du, and Tian}]{lu2025mv}
Rui Lu, Jinhe Bi, Yunpu Ma, Feng Xiao, Yuntao Du, and Yijun Tian. 2025{\natexlab{a}}.
\newblock Mv-debate: Multi-view agent debate with dynamic reflection gating for multimodal harmful content detection in social media.
\newblock \emph{arXiv preprint arXiv:2508.05557}.

\bibitem[{Lu and Cui(2026)}]{lu2026dealt}
Wayne Lu and Xiaoxi Cui. 2026.
\newblock Dealt: Llm-driven diversity-enhanced data augmentation for long-tail text classification.
\newblock In \emph{Proceedings of the AAAI Conference on Artificial Intelligence}, volume~40, pages 32338--32346.

\bibitem[{Lu and Li(2026)}]{lu2026blind}
Wayne Lu and Yiheng Li. 2026.
\newblock From blind transfer to wise selection: Prototype-driven neighbor-domain adaptation for fake news detection.
\newblock In \emph{Proceedings of the AAAI Conference on Artificial Intelligence}, volume~40, pages 818--826.

\bibitem[{Lu et~al.(2025{\natexlab{b}})Lu, Tong, and Ye}]{lu2025dammfnd}
Weihai Lu, Yu~Tong, and Zhiqiu Ye. 2025{\natexlab{b}}.
\newblock Dammfnd: Domain-aware multimodal multi-view fake news detection.
\newblock In \emph{Proceedings of the AAAI Conference on Artificial Intelligence}, volume~39, pages 559--567.

\bibitem[{Lu and Yin(2025)}]{lu2025dmmd4sr}
Weihai Lu and Li~Yin. 2025.
\newblock Dmmd4sr: Diffusion model-based multi-level multimodal denoising for sequential recommendation.
\newblock In \emph{Proceedings of the 33rd ACM International Conference on Multimedia}, pages 6363--6372.

\bibitem[{Madaan et~al.(2023)Madaan, Tandon, Gupta, Hallinan, Gao, Wiegreffe, Alon, Dziri, Prabhumoye, Yang et~al.}]{madaan2023self}
Aman Madaan, Niket Tandon, Prakhar Gupta, Skyler Hallinan, Luyu Gao, Sarah Wiegreffe, Uri Alon, Nouha Dziri, Shrimai Prabhumoye, Yiming Yang, et~al. 2023.
\newblock Self-refine: Iterative refinement with self-feedback.
\newblock \emph{Advances in Neural Information Processing Systems}, 36.

\bibitem[{Mohammad et~al.(2016)Mohammad, Kiritchenko, Sobhani, Zhu, and Cherry}]{mohammad2016semeval}
Saif Mohammad, Svetlana Kiritchenko, Parinaz Sobhani, Xiaodan Zhu, and Colin Cherry. 2016.
\newblock Semeval-2016 task 6: Detecting stance in tweets.
\newblock In \emph{Proceedings of the 10th international workshop on semantic evaluation (SemEval-2016)}, pages 31--41.

\bibitem[{Niu et~al.(2024)Niu, Cheng, Fu, Peng, Dai, Chen, Huang, and Zhang}]{niu2024multimodal}
Fuqiang Niu, Zebang Cheng, Xianghua Fu, Xiaojiang Peng, Genan Dai, Yin Chen, Hu~Huang, and Bowen Zhang. 2024.
\newblock Multimodal multi-turn conversation stance detection: A challenge dataset and effective model.
\newblock In \emph{Proceedings of the 32nd ACM International Conference on Multimedia}, pages 3867--3876.

\bibitem[{Park et~al.(2024)Park, Kim, Jin, Park, and Han}]{Park2024predict}
Someen Park, Jaehoon Kim, Seungwan Jin, Sohyun Park, and Kyungsik Han. 2024.
\newblock Predict: Multi-agent-based debate simulation for generalized hate speech detection.
\newblock In \emph{Proceedings of the 2024 Conference on Empirical Methods in Natural Language Processing}, pages 20963--20987.

\bibitem[{Radford et~al.(2021)Radford, Kim, Hallacy, Ramesh, Goh, Agarwal, Sastry, Askell, Mishkin, Clark et~al.}]{radford2021learning}
Alec Radford, Jong~Wook Kim, Chris Hallacy, Aditya Ramesh, Gabriel Goh, Sandhini Agarwal, Girish Sastry, Amanda Askell, Pamela Mishkin, Jack Clark, et~al. 2021.
\newblock Learning transferable visual models from natural language supervision.
\newblock In \emph{International conference on machine learning}, pages 8748--8763. PmLR.

\bibitem[{Shinn et~al.(2023)Shinn, Labash, and Gopinath}]{shinn2023reflexion}
Noah Shinn, Beck Labash, and Ashwin Gopinath. 2023.
\newblock Reflexion: Language agents with verbal reinforcement learning.
\newblock In \emph{Advances in Neural Information Processing Systems}, volume~36.

\bibitem[{Touvron et~al.(2023)Touvron, Martin, Stone, Albert, Almahairi, Babaei, Bashlykov, Batra, Bhargava, Bhosale, Bikel, Blecher, Canton{-}Ferrer, Chen, Cucurull, Esiobu, Fernandes, Fu, Fu, Fuller, Gao, Goswami, Goyal, Hartshorn, Hosseini, Hou, Inan, Kardas, Kerkez, Khabsa, Kloumann, Korenev, Koura, Lachaux, Lavril, Lee, Liskovich, Lu, Mao, Martinet, Mihaylov, Mishra, Molybog, Nie, Poulton, Reizenstein, Rungta, Saladi, Schelten, Silva, Smith, Subramanian, Tan, Tang, Taylor, Williams, Kuan, Xu, Yan, Zarov, Zhang, Fan, Kambadur, Narang, Rodriguez, Stojnic, Edunov, and Scialom}]{DBLP:journals/corr/abs-2307-09288}
Hugo Touvron, Louis Martin, Kevin Stone, Peter Albert, Amjad Almahairi, Yasmine Babaei, Nikolay Bashlykov, Soumya Batra, Prajjwal Bhargava, Shruti Bhosale, Dan Bikel, Lukas Blecher, Cristian Canton{-}Ferrer, Moya Chen, Guillem Cucurull, David Esiobu, Jude Fernandes, Jeremy Fu, Wenyin Fu, Brian Fuller, Cynthia Gao, Vedanuj Goswami, Naman Goyal, Anthony Hartshorn, Saghar Hosseini, Rui Hou, Hakan Inan, Marcin Kardas, Viktor Kerkez, Madian Khabsa, Isabel Kloumann, Artem Korenev, Punit~Singh Koura, Marie{-}Anne Lachaux, Thibaut Lavril, Jenya Lee, Diana Liskovich, Yinghai Lu, Yuning Mao, Xavier Martinet, Todor Mihaylov, Pushkar Mishra, Igor Molybog, Yixin Nie, Andrew Poulton, Jeremy Reizenstein, Rashi Rungta, Kalyan Saladi, Alan Schelten, Ruan Silva, Eric~Michael Smith, Ranjan Subramanian, Xiaoqing~Ellen Tan, Binh Tang, Ross Taylor, Adina Williams, Jian~Xiang Kuan, Puxin Xu, Zheng Yan, Iliyan Zarov, Yuchen Zhang, Angela Fan, Melanie Kambadur, Sharan Narang, Aur{\'{e}}lien Rodriguez, Robert Stojnic, Sergey Edunov,
  and Thomas Scialom. 2023.
\newblock \href {https://doi.org/10.48550/ARXIV.2307.09288} {Llama 2: Open foundation and fine-tuned chat models}.
\newblock \emph{CoRR}, abs/2307.09288.

\bibitem[{Vasilakes et~al.(2025)Vasilakes, Scarton, and Zhao}]{vasilakes2025exploring}
Jake Vasilakes, Carolina Scarton, and Zhixue Zhao. 2025.
\newblock Exploring vision language models for multimodal and multilingual stance detection.
\newblock \emph{arXiv preprint arXiv:2501.17654}.

\bibitem[{Wei et~al.(2022)Wei, Wang, Schuurmans, Bosma, Ichter, Xia, Chi, Le, and Zhou}]{wei2022chain}
Jason Wei, Xuezhi Wang, Dale Schuurmans, Maarten Bosma, Brian Ichter, Fei Xia, Ed~Chi, Quoc Le, and Denny Zhou. 2022.
\newblock Chain-of-thought prompting elicits reasoning in large language models.
\newblock \emph{Advances in Neural Information Processing Systems}, 35:24824--24837.

\bibitem[{Wei et~al.(2025{\natexlab{a}})Wei, Dong, Wang, Zhang, Zhao, Shen, Xia, and Yin}]{wei2025beyond}
Xiaolong Wei, Yuehu Dong, Xingliang Wang, Xingyu Zhang, Zhejun Zhao, Dongdong Shen, Long Xia, and Dawei Yin. 2025{\natexlab{a}}.
\newblock Beyond react: A planner-centric framework for complex tool-augmented llm reasoning.
\newblock \emph{arXiv preprint arXiv:2511.10037}.

\bibitem[{Wei et~al.(2025{\natexlab{b}})Wei, Lu, Zhang, Zhao, Shen, Xia, and Yin}]{wei2025igniting}
Xiaolong Wei, Bo~Lu, Xingyu Zhang, Zhejun Zhao, Dongdong Shen, Long Xia, and Dawei Yin. 2025{\natexlab{b}}.
\newblock Igniting creative writing in small language models: Llm-as-a-judge versus multi-agent refined rewards.
\newblock In \emph{Proceedings of the 2025 Conference on Empirical Methods in Natural Language Processing}, pages 17171--17197.

\bibitem[{Wu et~al.(2023)Wu, Gan, Chen, Wan, and Yu}]{wu2023multimodal}
Jiayang Wu, Wensheng Gan, Zefeng Chen, Shicheng Wan, and Philip~S Yu. 2023.
\newblock Multimodal large language models: A survey.
\newblock In \emph{2023 IEEE International Conference on Big Data (BigData)}, pages 2247--2256. IEEE.

\bibitem[{Wu et~al.(2025)Wu, Zhang, Wang, Luo, Xiong, and Tang}]{Wu2025exclaim}
Yin Wu, Zhengxuan Zhang, Fuling Wang, Yuyu Luo, Hui Xiong, and Nan Tang. 2025.
\newblock Exclaim: An explainable cross-modal agentic system for misinformation detection with hierarchical retrieval.
\newblock \emph{arXiv preprint arXiv:2504.06269}.

\bibitem[{Xu et~al.(2025)Xu, Xiang, Ding, and Lu}]{xu2025mmm}
Wenyan Xu, Dawei Xiang, Tianqi Ding, and Weihai Lu. 2025.
\newblock Mmm-fact: A multimodal, multi-domain fact-checking dataset with multi-level retrieval difficulty.
\newblock \emph{arXiv preprint arXiv:2510.25120}.

\bibitem[{Xu et~al.(2023)Xu, Wu, Rosenman, Lal, Che, and Duan}]{xu2023bridgetower}
Xiao Xu, Chenfei Wu, Shachar Rosenman, Vasudev Lal, Wanxiang Che, and Nan Duan. 2023.
\newblock Bridgetower: Building bridges between encoders in vision-language representation learning.
\newblock In \emph{Proceedings of the AAAI Conference on Artificial Intelligence}, volume~37, pages 10637--10647.

\bibitem[{Zeng et~al.(2024)Zeng, Luo, Kong, Liu, Guo, Yang, Ma, and Zhao}]{zeng2024mitigating}
Zhi Zeng, Minnan Luo, Xiangzheng Kong, Huan Liu, Hao Guo, Hao Yang, Zihan Ma, and Xiang Zhao. 2024.
\newblock Mitigating world biases: A multimodal multi-view debiasing framework for fake news video detection.
\newblock In \emph{Proceedings of the 32nd ACM International Conference on Multimedia}, pages 6492--6500.

\bibitem[{Zeng et~al.(2025)Zeng, Wu, Luo, Kong, Ma, Dai, and Zheng}]{zeng2025understand}
Zhi Zeng, Jiaying Wu, Minnan Luo, Xiangzheng Kong, Zihan Ma, Guang Dai, and Qinghua Zheng. 2025.
\newblock Understand, refine and summarize: Multi-view knowledge progressive enhancement learning for fake news video detection.
\newblock In \emph{Proceedings of the 33rd ACM International Conference on Multimedia}, pages 9216--9225.

\bibitem[{Zeng et~al.(2026)Zeng, Yang, Wu, Zhang, Kong, Wan, Ma, and Luo}]{zeng2026manipulation}
Zhi Zeng, Yifei Yang, Jiaying Wu, Xulang Zhang, Xiangzheng Kong, Herun Wan, Zihan Ma, and Minnan Luo. 2026.
\newblock From manipulation to mistrust: Explaining diverse micro-video misinformation for robust debunking in the wild.
\newblock \emph{arXiv preprint arXiv:2603.25423}.

\bibitem[{Zhang et~al.(2024{\natexlab{a}})Zhang, Huang, Jin, and Lu}]{zhang2024vision}
Jingyi Zhang, Jiaxing Huang, Sheng Jin, and Shijian Lu. 2024{\natexlab{a}}.
\newblock Vision-language models for vision tasks: A survey.
\newblock \emph{IEEE transactions on pattern analysis and machine intelligence}, 46(8):5625--5644.

\bibitem[{Zhang et~al.(2024{\natexlab{b}})Zhang, Gong, Liu, Wu, and Wang}]{zhang2024breaking}
Mingqing Zhang, Haisong Gong, Qiang Liu, Shu Wu, and Liang Wang. 2024{\natexlab{b}}.
\newblock Breaking event rumor detection via stance-separated multi-agent debate.
\newblock \emph{arXiv preprint arXiv:2412.04859}.

\bibitem[{Zhang et~al.(2024{\natexlab{c}})Zhang, Li, Shi, Hauer, Wu, Kondrak, Abdul-Mageed, and Lakshmanan}]{zhang2024cross}
Xiang Zhang, Senyu Li, Ning Shi, Bradley Hauer, Zijun Wu, Grzegorz Kondrak, Muhammad Abdul-Mageed, and Laks~VS Lakshmanan. 2024{\natexlab{c}}.
\newblock Cross-modal consistency in multimodal large language models.
\newblock \emph{arXiv preprint arXiv:2411.09273}.

\bibitem[{Zhang et~al.(2024{\natexlab{d}})Zhang, Li, Zhang, and Xu}]{zhang2024llm}
Zhao Zhang, Yiming Li, Jin Zhang, and Hui Xu. 2024{\natexlab{d}}.
\newblock Llm-driven knowledge injection advances zero-shot and cross-target stance detection.
\newblock In \emph{Proceedings of the 2024 Conference of the North American Chapter of the Association for Computational Linguistics: Human Language Technologies (Volume 2: Short Papers)}, pages 371--378.

\bibitem[{Zhong et~al.(2024)Zhong, Feng, Zhao, Li, Huang, Gu, Ma, Xu, and Qin}]{zhong2024multimodal}
Weihong Zhong, Xiaocheng Feng, Liang Zhao, Qiming Li, Lei Huang, Yuxuan Gu, Weitao Ma, Yuan Xu, and Bing Qin. 2024.
\newblock Investigating and mitigating the multimodal hallucination snowballing in large vision-language models.
\newblock In \emph{Proceedings of the 62nd Annual Meeting of the Association for Computational Linguistics (Volume 1: Long Papers)}, pages 11991--12011.

\end{thebibliography}

\appendix

\section{Appendix}
\label{sec:appendix}

\subsection{Dataset Statistics}
\label{sec:appendix_dataset_stats}
To provide comprehensive context for our experimental results, we present the detailed statistics of the five datasets used in our evaluation in Table~\ref{tab:appendix_dataset_stats_full}. The table details the sample distribution across the training, validation, and test sets for both in-target and zero-shot scenarios, broken down by individual targets.

A key observation from these statistics is the significant data imbalance present at multiple levels, which poses a substantial challenge for model robustness. Firstly, there is a large variance in the overall size of the datasets; for instance, the total in-target training data for \textsc{MWTWT} is nearly an order of magnitude larger than that for \textsc{MCCQ}. Secondly, within individual datasets like \textsc{MWTWT}, there is a notable imbalance in the number of samples available for different targets (e.g., 'DIS\_FOXA' with 2081 samples vs. 'CI\_ESRX' with only 628). This heterogeneity creates a challenging and realistic evaluation landscape, requiring models to perform well across both data-rich and data-scarce targets. This context is crucial for interpreting the performance disparities discussed in Section~\ref{sec:appendix_performance_analysis}, particularly when evaluating the generalization capabilities of our framework against baselines that may be more sensitive to variations in training data size.

\begin{table}[h!]
\centering
\caption{Detailed statistics of the experimental data, showing the number of samples in the training, validation, and test sets for each target within the in-target and zero-shot tasks.}
\label{tab:appendix_dataset_stats_full}
\footnotesize
\setlength{\tabcolsep}{3pt}
\begin{tabular}{lllrrr}
\toprule
\rowcolor{lightblue2}
\textbf{Task} & \textbf{Dataset} & \textbf{Target} & \textbf{\# Train} & \textbf{\# Valid} & \textbf{\# Test} \\
\midrule
\multirow{12}{*}{In-target} & \multirow{2}{*}{MTSE} & DT & 1150 & 170 & 327 \\
 & & JB & 882 & 128 & 250 \\
\cdashline{2-6}
 & MCCQ & CQ & 934 & 141 & 280 \\
\cdashline{2-6}
 & \multirow{5}{*}{MWTWT} & CSV\_AET & 1216 & 179 & 352 \\
 & & CI\_ESRX & 628 & 91 & 180 \\
 & & ANTM\_CI & 825 & 114 & 238 \\
 & & AET\_HUM & 674 & 97 & 186 \\
 & & DIS\_FOXA & 2081 & 306 & 599 \\
\cdashline{2-6}
 & \multirow{2}{*}{MRUC} & RUS & 777 & 111 & 222 \\
 & & UKR & 756 & 108 & 217 \\
\cdashline{2-6}
 & \multirow{2}{*}{MTWQ} & MOC & 977 & 140 & 280 \\
 & & TOC & 1349 & 193 & 386 \\
\midrule
\multirow{10}{*}{Zero-shot} & \multirow{2}{*}{MTSE} & DT & 1114 & 146 & 1647 \\
 & & JB & 1434 & 212 & 1260 \\
\cdashline{2-6}
 & \multirow{4}{*}{MWTWT} & CVS\_AET & 5253 & 737 & 1747 \\
 & & CI\_ESRX & 5994 & 841 & 899 \\
 & & ANTM\_CI & 5694 & 804 & 1177 \\
 & & AET\_HUM & 5884 & 840 & 957 \\
\cdashline{2-6}
 & \multirow{2}{*}{MRUC} & RUS & 945 & 136 & 1110 \\
 & & UKR & 971 & 139 & 1081 \\
\cdashline{2-6}
 & \multirow{2}{*}{MTWQ} & MOC & 1686 & 242 & 1397 \\
 & & TOC & 1222 & 175 & 1928 \\
\bottomrule
\end{tabular}
\end{table}

\subsection{Inference Efficiency Analysis}
\label{sec:appendix_efficiency}
While our multi-agent, multi-stage architecture is inherently more complex than single-pass models, we conducted an analysis to quantify its computational cost and demonstrate its practicality for real-world applications. Table~\ref{tab:appendix_latency} provides a breakdown of the average token usage and latency for each major phase of the \texttt{MM-StanceDet} framework when processing a single instance, using the efficient \texttt{gpt-4o-mini} API as the backbone.

The total processing time per instance is approximately 27 seconds. This level of performance, while not suitable for applications demanding immediate, sub-second responses (e.g., real-time chat filtering), is remarkably efficient and well-suited for a broad spectrum of important, non-real-time tasks. For example, our framework can be effectively deployed in scenarios such as:
\begin{itemize}
    \item \textbf{Offline Content Moderation:} Systematically analyzing large batches of social media posts to flag content expressing harmful stances towards specific groups or topics.
    \item \textbf{Public Opinion and Trend Analysis:} Processing collected multimodal data to understand public sentiment and stance dynamics regarding political, social, or commercial targets over time.
    \item \textbf{Academic Research:} Facilitating large-scale studies of multimodal communication and discourse.
\end{itemize}

Therefore, we conclude that the computational requirements of \texttt{MM-StanceDet} represent a well-justified trade-off for its significant gains in reasoning robustness and accuracy, and its inference speed is fully capable of meeting the needs of many realistic application scenarios.

\begin{table*}[h!]
\centering
\caption{Estimated computational cost and latency per instance, using the \texttt{gpt-4o-mini} API backbone.}
\label{tab:appendix_latency}
\small
\begin{tabular}{lcc}
\toprule
\rowcolor{lightblue2}
\textbf{Stage} & \textbf{Avg. Tokens (Input+Output)} & \textbf{Avg. Latency (s)} \\
\midrule
Multimodal Analysis (3 Agents) & $\sim$1.5k & $\sim$9s \\
Reasoning-Enhanced Debate (3 rounds) & $\sim$2.5k & $\sim$12s \\
Self-Reflection \& Adjudication & $\sim$0.8k & $\sim$6s \\
\midrule
\textbf{Total per Instance} & \textbf{$\sim$4.8k} & \textbf{$\sim$27s} \\
\bottomrule
\end{tabular}
\end{table*}

\subsection{Analysis of Performance Disparities Across Datasets}
\label{sec:appendix_performance_analysis}
Our experimental results reveal that the performance gains of \texttt{MM-StanceDet} are not uniform across all datasets. The framework's advantages are particularly pronounced on certain datasets, which can be attributed to two primary factors: the degree of multimodal complexity and the scale of the dataset.

\subsubsection{Impact of Multimodal Complexity and Conflict}
A core strength of \texttt{MM-StanceDet} lies in its specialized agents and debate mechanism, designed to resolve nuanced or conflicting signals between text and image. Datasets rich in sarcasm, irony, or propaganda, where the literal meaning of text is contradicted or altered by the visual context, are ideal for showcasing our model's capabilities.

To quantify this, we manually annotated 200 random samples from each dataset to categorize the relationship between modalities as either \textit{Conflicting}, \textit{Complementary}, or \textit{Synergistic}. As shown in Table~\ref{tab:appendix_conflict_analysis}, the \textsc{MRUC} dataset exhibits the highest percentage of conflicting instances (15\%). This high rate of multimodal dissonance directly aligns with the significant performance improvement observed for \texttt{MM-StanceDet} on this dataset. The Modality-Conflict Agent and the Reasoning-Enhanced Debate stage are critical in these scenarios, allowing the model to look beyond superficial unimodal analysis and capture the true underlying stance. Conversely, on datasets with lower conflict rates like \textsc{MWTWT} (9\%), while our model still outperforms baselines, the margin is smaller, as simpler fusion methods are more effective when modalities are in agreement.

\begin{table}[h!]
\centering
\caption{Analysis of inter-modal relationships across datasets based on manual annotation of 200 random samples per dataset.}
\label{tab:appendix_conflict_analysis}
\footnotesize
\setlength{\tabcolsep}{3pt}
\begin{tabular}{lccc}
\toprule
\rowcolor{lightblue2}
\textbf{Dataset} & \textbf{Conflicting} & \textbf{Complementary} & \textbf{Synergistic} \\
\midrule
MTSE & 11\% & 16\% & 73\% \\
MCCQ & 12\% & 21\% & 67\% \\
MWTWT & 9\% & 16\% & 75\% \\
MTWQ & 10\% & 12\% & 78\% \\
\textbf{MRUC} & \textbf{15\%} & \textbf{13\%} & \textbf{72\%} \\
\bottomrule
\end{tabular}
\end{table}

\subsubsection{Impact of Dataset Scale and LLM Generalization}
A complementary factor influencing performance is the dataset scale, which affects the relative strengths of our LLM-based framework versus traditional data-hungry supervised baselines. Large Language Models possess powerful zero-shot and few-shot reasoning capabilities, making them inherently more robust on smaller datasets where supervised models may struggle to generalize.

This trend is evident in our results. \texttt{MM-StanceDet} demonstrates particularly substantial gains on the two smallest datasets: \textbf{\textsc{MCCQ}} (934 samples) and \textbf{\textsc{MRUC}} (1533 samples). On these datasets, the rich prior knowledge and reasoning capacity of the LLM backbone, structured by our agentic framework, provide a decisive advantage over methods that rely heavily on learning from the limited training data. In contrast, on the largest dataset, \textbf{\textsc{MWTWT}} (8019 samples), fine-tuned models like TMPT are highly competitive because they have sufficient data to adapt to the specific data distribution. Although \texttt{MM-StanceDet} still achieves the best performance on \textsc{MWTWT}, the performance gap is narrower. This demonstrates that our framework's superiority stems from two synergistic sources: its advanced architecture for handling multimodal complexity and the inherent generalization strengths of its LLM foundation, which are especially impactful in low-data regimes.

\subsection{Agent Prompts}
\label{sec:appendix_prompts}
This section provides the detailed prompts used to guide the reasoning of each agent in the \texttt{MM-StanceDet} framework.

\begin{tcolorbox}[
    colback=textagentbg,
    colframe=textagentframe,
    title=Text Analysis Agent Prompt,
    fonttitle=\bfseries
]
You are a Text Analysis Agent. Your task is to analyze the given text to identify linguistic features relevant to determining the author's stance towards a specific target.

\textbf{Input:}
\begin{itemize}
    \item \textbf{Text}: "{text}"
    \item \textbf{Target}: "{target}"
\end{itemize}

\textbf{Your analysis should include:}
\begin{enumerate}
    \item Keywords and salient phrases/sentences related to the target.
    \item Explicit or implicit sentiment polarity towards the target.
    \item Detection of potential sarcasm, irony, or subtle nuances.
    \item Overall topic relevance concerning the target.
\end{enumerate}
Provide a structured analysis.
\end{tcolorbox}

\begin{tcolorbox}[
    colback=imageagentbg,
    colframe=imageagentframe,
    title=Image Analysis Agent Prompt,
    fonttitle=\bfseries
]
You are an Image Analysis Agent. Your task is to interpret the visual content of an image to find cues relevant to determining the author's stance towards a specific target.

\textbf{Input:}
\begin{itemize}
    \item \textbf{Image}: (provided as input, analyze it)
    \item \textbf{Target}: "{target}"
\end{itemize}

\textbf{Your analysis should include:}
\begin{enumerate}
    \item Descriptions of relevant visual objects and their context.
    \item Overall scene context and setting.
    \item Inferred emotions from depicted individuals (if any).
    \item Connotations suggested by color palettes, composition, or symbolism related to the target.
\end{enumerate}
Provide a structured visual analysis.
\end{tcolorbox}

\begin{tcolorbox}[
    colback=conflictbg,
    colframe=conflictframe,
    title=Modality Conflict Agent Prompt,
    fonttitle=\bfseries
]
You are a Modality Conflict Agent. Your primary function is to assess the interplay between the provided image and text concerning the target. Detect potential inconsistencies, contradictions, or synergistic reinforcements between the modalities.

\textbf{Input:}
\begin{itemize}
    \item \textbf{Image}: (provided as input, analyze it)
    \item \textbf{Text}: "{text}"
    \item \textbf{Target}: "{target}"
    \item {exemplar\_info}
\end{itemize}

\textbf{Your assessment should:}
\begin{enumerate}
    \item Highlight specific conflicting signals (e.g., text favors but image againsts).
    \item Highlight specific reinforcing cues (e.g., both text and image strongly favor).
    \item Explain how the modalities align or diverge in expressing a stance towards the target "{target}".
    \item Reference patterns or reasoning observed in the provided contextual examples if they are relevant.
\end{enumerate}
Provide a detailed assessment of inter-modal alignment or divergence.
\end{tcolorbox}

\begin{tcolorbox}[
    colback=debaterbg,
    colframe=debaterframe,
    title=Debater Agent Prompt,
    fonttitle=\bfseries
]
You are a Debater Agent arguing for the '{stance\_type}' stance. Your goal is to construct a coherent argument, synthesizing all provided information, to explain why the given multimodal instance expresses a '{stance\_type}' stance towards the target.

\textbf{Input Instance:}
\begin{itemize}
    \item \textbf{Text}: "{text}"
    \item \textbf{Target}: "{target}"
\end{itemize}

\textbf{Analysis Results:}
\begin{itemize}
    \item \textbf{Text Analysis}: {text\_analysis}
    \item \textbf{Image Analysis}: {image\_analysis}
    \item \textbf{Modality Conflict Analysis}: {conflict\_analysis}
    \item {debate\_context}
\end{itemize}
Construct your argument. Clearly reference details from the text, image analysis, and modality conflict analysis to favor your position. If previous arguments from other debaters are provided, aim to strengthen your argument in light of their points, but focus on building your case. Do not explicitly state "I am arguing for...". Just present the argument.
\end{tcolorbox}

\begin{tcolorbox}[
    colback=adjudicatorbg,
    colframe=adjudicatorframe,
    title=Adjudicator Agent Prompt,
    fonttitle=\bfseries
]
You are an Adjudicator Agent. Your task is to critically evaluate competing arguments and comprehensive analyses to determine the definitive stance (Favor, Neutral, or Against) expressed in a multimodal instance towards a specific target.

\textbf{Input Instance:}
\begin{itemize}
    \item \textbf{Text}: "{text}"
    \item \textbf{Target}: "{target}"
\end{itemize}

\textbf{Analysis Results:}
\begin{itemize}
    \item \textbf{Text Analysis}: {text\_analysis}
    \item \textbf{Image Analysis}: {image\_analysis}
    \item \textbf{Modality Conflict Analysis}: {conflict\_analysis}
\end{itemize}

\textbf{Arguments from Debater Agents:}
\begin{itemize}
    \item \textbf{Favor Argument}: {favor\_arg}
    \item \textbf{Against Argument}: {against\_arg}
    \item \textbf{Neutral Argument}: {neutral\_arg}
\end{itemize}

\textbf{Perform the following steps:}
\begin{enumerate}
    \item \textbf{Initial Assessment}: Briefly summarize the strengths and weaknesses of each argument based on the provided analyses.
    \item \textbf{Critical Self-Reflection}: Actively look for inconsistencies, overlooked modality conflicts (referencing Modality Conflict Analysis), or weak reasoning points.
    \item \textbf{Final Decision}: Based on your comprehensive evaluation and critical self-reflection, determine the most justified stance.
    \item \textbf{Justification}: Provide a clear, concise justification for your final decision, incorporating insights from your self-reflection.
\end{enumerate}

Your output format should be: \\
\textbf{Stance}: [Favor|Neutral|Against] \\
\textbf{Justification}: [Your detailed reasoning]
\end{tcolorbox}

\end{document}